\documentclass[10pt,twocolumn,letterpaper]{article}

\usepackage{wacv}

\makeatletter
\@namedef{ver@everyshi.sty}{}
\makeatother
\usepackage{tikz}

\usepackage{times}
\usepackage{epsfig}
\usepackage{graphicx}
\usepackage{amsmath}
\usepackage{amssymb}
\usepackage{xcolor}
\usepackage[linesnumbered,ruled,vlined]{algorithm2e}
\usepackage{tabularx}
\usepackage{booktabs}
\usepackage{enumitem}
\usepackage[misc]{ifsym}
\usepackage[accsupp]{axessibility}

\SetCommentSty{mycommfont}

\SetKwInput{KwInput}{Input}                
\SetKwInput{KwOutput}{Output}              

\DeclareMathOperator*{\argmax}{arg\,max}

\newcommand\copyrighttext{%
  \centering    
  \footnotesize ~\copyright 20xx IEEE. To appear, Proceedings of the 2022 IEEE/CVF Winter Conference on Applications of Computer Vision (WACV)}
\newcommand\copyrightnotice{%
\begin{tikzpicture}[remember picture,overlay]
\node[anchor=south,yshift=10pt] at (current page.south) {\fbox{\parbox{\dimexpr\textwidth-\fboxsep-\fboxrule\relax}{\copyrighttext}}};
\end{tikzpicture}%
}

\def\wacvPaperID{800} 

\wacvfinalcopy 

\ifwacvfinal
\def\assignedStartPage{9876} 
\fi

\ifwacvfinal
\usepackage[breaklinks=true,bookmarks=false]{hyperref}
\else
\usepackage[pagebackref=true,breaklinks=true,colorlinks,bookmarks=false]{hyperref}
\fi

\ifwacvfinal
\else
\fi

\def\httilde{\mbox{\tt\raisebox{-.5ex}{\symbol{126}}}}

\begin{document}

\title{Bayesian Uncertainty and Expected Gradient Length - Regression:\\ Two Sides Of The Same Coin?}

\author{Megh Shukla \\
Mercedes-Benz Research and Development India \\
{\tt\small megh.shukla@daimler.com}

}

\maketitle

\begin{abstract}
    Active learning algorithms select a subset of data for annotation to maximize the model performance on a budget. One such algorithm is Expected Gradient Length, which as the name suggests uses the approximate gradient induced per example in the sampling process. While Expected Gradient Length has been successfully used for classification and regression, the formulation for regression remains intuitively driven. Hence, our theoretical contribution involves deriving this formulation, thereby supporting experimental evidence \cite{batch_egl_regression, egl_regression}. Subsequently, we show that expected gradient length in regression is equivalent to Bayesian uncertainty \cite{kendall2017uncertainties}. If certain assumptions are infeasible, our algorithmic contribution (EGL++) approximates the effect of ensembles with a single deterministic network. Instead of computing multiple possible inferences per input, we leverage previously annotated samples to quantify the probability of previous labels being the true label. Such an approach allows us to extend expected gradient length to a new task: human pose estimation. We perform experimental validation on two human pose datasets (MPII and LSP/LSPET), highlighting the interpretability and competitiveness of EGL++ with different active learning algorithms for human pose estimation.
    \copyrightnotice
    
\end{abstract}

\section{Introduction}

\begin{figure}[th!]
   \centering
   \includegraphics[width=0.9\linewidth]{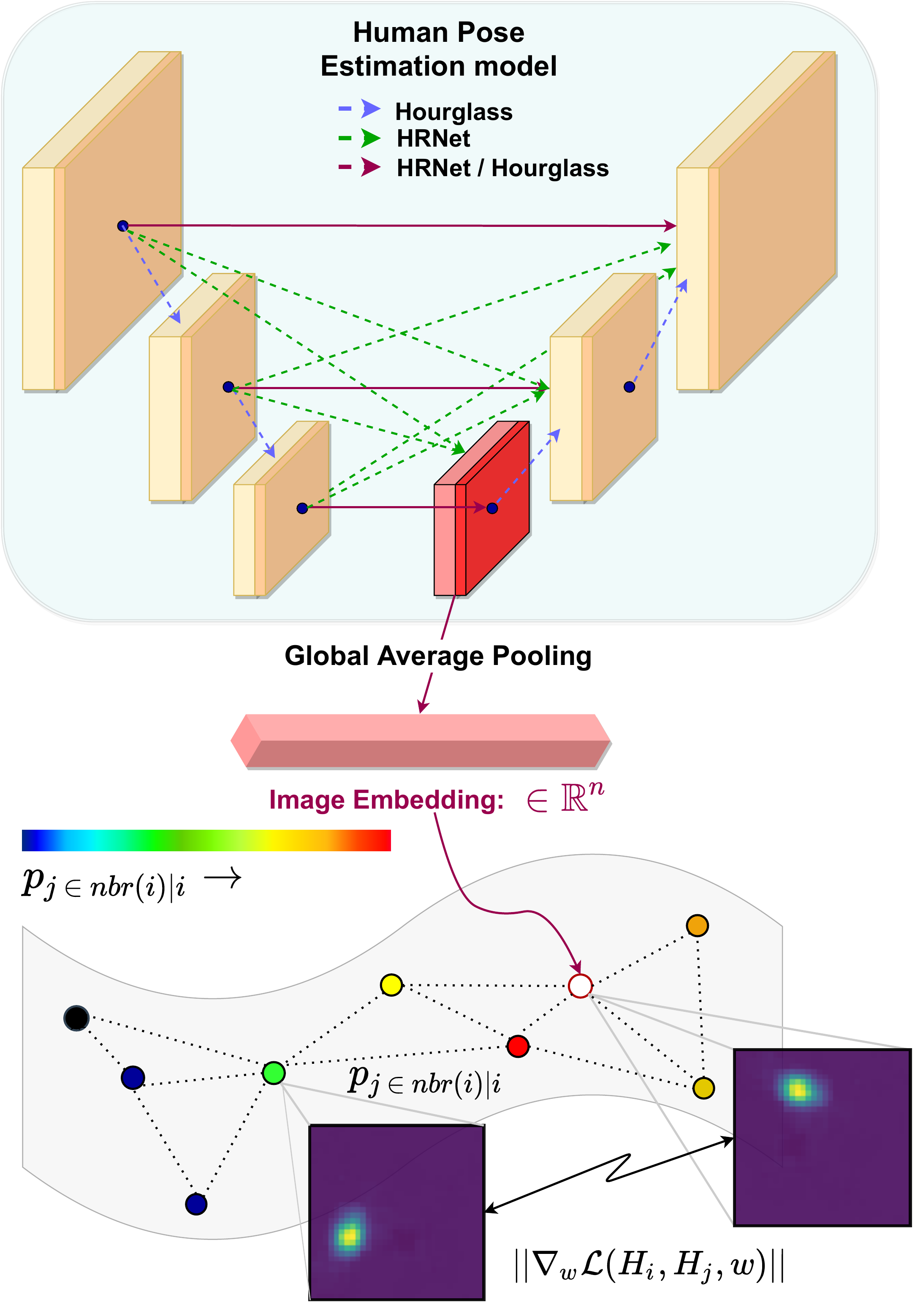}
   \caption{While our theoretical contribution links expected gradient length (EGL) to Bayesian uncertainty, our algorithmic contribution approximates the effect of ensembles with a single deterministic network: The notion of neighbors for an (image, target) pair uses low-dimensional representations from the neural network. These representations allow us to interpret the embedding space into neighborhood distributions \cite{tsne}. This neighborhood distribution quantifies the probability of various inferences being the target inference for the given sample, approximating an ensemble. Not only do we leverage previously labelled samples within the EGL framework, the approach quantifies the intuition that similar inputs have similar representation. Being label shape agnostic, EGL++ extends EGL to the new task of human pose estimation.}
\label{fig:egl}
\end{figure}

Imagine that as researchers, we are tasked with improving computer vision models for a client's products, deployed across the globe. The challenge is that training datasets may not reflect all real world use cases. While we could improve real-world performance by collecting data from end-users, the resultant collection would be humongous and data annotation is arduous. \textit{Instead, can we identify a subset of images for annotation, to maximize the model performance per set of images annotated? Simultaneously, can we improve the interpretability in our sampling process?}

\par     
Our search for a solution leads us to Active Learning (AL) \cite{settles2009active}, a suite of algorithms designed to select a subset of images for annotation when labelling the entire unlabelled pool of data is infeasible. This subset is obtained using the model's feedback, ensuring that the selected images impart new information for the model to learn. AL is a cyclical process of selection-annotation-training, allowing for fast prototyping, lower budget requirements and potentially limiting bias in the data. AL approaches span uncertainty \cite{gawlikowski2021survey}, ensemble \cite{melville2004diverse} and diversity \cite{sener2017active} to name a few. While diversity based approaches seek to introduce semantic variations, uncertainty quantifies model ambiguity to select images for annotation. Ensemble approaches personify wisdom of the masses, with multiple models used to determine the annotation set.
\par
We focus on Expected Gradient Length (EGL) \cite{settles2009active}, a classical AL algorithm used in diverse applications such as speech recognition, text representation and biomedical imagery. While EGL has been extensively studied in classification, recent approaches apply the framework for regression \cite{batch_egl_regression, egl_regression}. However, the formulation for regression remains intuitively driven and lacks theoretical background. With this motivation, we focus on supporting the empirical results in \cite{batch_egl_regression, egl_regression} by developing theory based on previous statistical results involving Fisher information \cite{huang2016active, ly2017tutorial, settles2008analysis, sourati2017asymptotic}. Additionally, we also explore an alternative interpretation of the framework, allowing us to extend EGL to the new task of human pose estimation. Our contributions are:
\setlist{nolistsep}
\begin{enumerate}[noitemsep]
    \item \textbf{Theory}: We theoretically derive the formulation for EGL - regression \cite{batch_egl_regression, egl_regression}, and further show the equivalency between the resultant closed form solution and predictive variance \cite{kendall2017uncertainties}.
    
    \item \textbf{Algorithm}: EGL++ is an efficient, alternative interpretation of the EGL framework. We show that the effect of an ensemble can be approximated with a single deterministic network by leveraging  previously labelled data explicitly. This interpretation additionally allows us to extend expected gradient length to the completely new task of human pose estimation.  
\end{enumerate}
Similar inputs have similar representations since neural networks are almost everywhere differentiable. EGL++ quantifies this intuition, weighing the importance previously labelled data points in the neighborhood of any given sample have. Since this approach depends on neural representations and agnostic of the target inference shape, EGL++ allows us to extend the EGL paradigm to human pose estimation (HPE). Active learning for HPE is tricky \footnote{Elaborated in \textit{Relation to Prior Art}, Related Work section}; popular architectures for HPE are fully \textit{convolutional} and regress 2D heatmaps, limiting the choice of AL algorithms. We show that EGL++ is competitive in maximizing the performance of human pose models when full annotation is infeasible, with an added advantage of interpretability. 


\section{Related Work}

\textbf{Active Learning:} Settles' active learning survey \cite{settles2009active} remains a gold mine of information, detailing various classical methods. Classical uncertainty algorithms such as \cite{ent2, ent1} have a rich history, employing variations of entropy and maximum margin using softmax outputs of the model. Recent works include \cite{gan_multiclass} which use GANs to generate high entropy samples for multi-class problems, and \cite{uncertainty_segmentation} which employs uncertainty techniques for region based segmentation. 
\par
Ensembles \cite{beluch2018power, korner2006multi, melville2004diverse} build upon the Query by Committee paradigm, using multiple models to select the annotation set. \cite{melville2004diverse} builds ensembles with artificial training data, \cite{beluch2018power} compares ensemble learning with recent active learning techniques and \cite{korner2006multi} uses cues from the softmax outputs in classification to show the use of maximum margin in a multi-class setting.
Diversity algorithms \cite{div2, sener2017active, div1} as the name suggests seek to incorporate a high degree of variation in the sampled set of unlabelled images. Core-set \cite{sener2017active} remains a popular approach that utilizes the linearly separable embedding space in the penultimate layer of classification networks.
Uncertainties in deep learning is an exciting area of research, summarized in Gawlikowski \etal \cite{gawlikowski2021survey}. Bayesian techniques \cite{decomposition, gal2016dropout, gal2017deep, bacoun, kendall2017uncertainties, gp_heteroscedastic} provide theoretical results to quantify uncertainty with the network predictions. These approaches rely on estimating the aleatoric and epistemic uncertainties to quantify the ambiguity associated with data and network inferences. Amersfoort \etal \cite{van2020uncertainty} borrows ideas from RBF networks and computes uncertainty in a single forward pass. Other approaches include Learning Loss \cite{Shukla_2021_CVPR, learnloss} which use an auxiliary network to predict the 'loss' for an image.

\par
\textbf{Expected Gradient Length:} Expected gradient length (EGL) \cite{settles2009active} utilizes the gradient norm in determining the most informative samples. Early works \cite{settles2008analysis, NIPS2007_a1519de5, degal, yuan} successfully leveraged EGL in classification and text. Huang \etal \cite{huang2016active} show that expected gradient length is a consequence of reducing the variance of the estimator over the testing set. Zhang \etal \cite{zhang2017active} use expected gradient length for sentence and document classification using CNNs. While previous methods worked with discrete outputs, \cite{batch_egl_regression, egl_regression} demonstrates the use of expected gradient length in regression. 

\par
\textbf{Human Pose Estimation: } Single person human pose estimation (HPE) has been widely studied \cite{bulat2020toward, hg, hrnet, deeppose} in literature. Popular architectures (Fig: \ref{fig:egl}) draw inspiration from U-Net \cite{unet} extracting features from multiple scales using a top-down approach. Human pose models take as input 3-channel RGB images and regresses a  2-D heatmap (one for each joint), denoting the location for the joint. Active learning for human pose estimation was first discussed in \cite{liu2017active}, which proposed multi-peak entropy by computing, normalizing, and performing softmax over the local maxima present in the heatmaps.

\par
\textbf{Relation to Prior Art: } Our work is geared towards expected gradient length for regression and succeeds Cai \etal \cite{batch_egl_regression, egl_regression}. We theoretically derive the formulation, supporting the experimental results in Cai \etal. We go one step further and show that expected gradient length in regression is equivalent to prediction uncertainty. Our algorithmic contribution (EGL++) approximates ensembles with a single deterministic network by utilizing previously labelled samples within the EGL framework. Both DUQ \cite{van2020uncertainty} and EGL++ use distances as a measure of uncertainty. However, DUQ learns class specific weight matrices to encode features in classification, whereas EGL++ uses learned representations directly in t-SNE. Gradient penalty in DUQ serves as a regularizer to prevent feature collapse, whereas gradients form the core of EGL++. \textit{(Details in the supplementary material)}

\par
\phantomsection
\label{par:challenge}
\textbf{Challenges with Human Pose: }While previous literature explores EGL for classification and regression, we extend the EGL framework to include human pose estimation (HPE). Our contribution is non-trivial since the challenges posed by HPE, namely a fully convolutional architecture regressing 2D heatmaps limits the choice of active learning algorithms. Bayesian uncertainty uses dropouts, causing convolutional architectures to suffer from strong regularization \cite{handpose}. Entropy based approaches are shown to be less effective \cite{learnloss}. HPE models are not deployed as ensembles and do not use voting. Core-set uses the fact that classification enforces linear separability in the penultimate layer \cite{sener2017active}, which does not hold true for HPE. Learning loss \cite{Shukla_2021_CVPR, learnloss} specializes in detecting \textit{lossy} images and is not suitable for generalized active learning. Some close works include bayesian hand pose estimation \cite{handpose}, however the network architecture and inputs differ significantly from human pose estimation. Aleatoric uncertainty has been effectively used in body joint occlusions \cite{ajain}, however the algorithm does not cater to active learning. Concerns \cite{sener2017active, learnloss} also remain on scalability of bayesian methods to large datasets.

\subsection{Revisiting Expected Gradient Length}

The intuition supporting Expected Gradient Length (EGL) is simple: using gradient as a measure of change imparted in the model by a given sample. While a converged network incurs negligible gradient across the training set, a sample from a different distribution incurs a large gradient since it imparts new information to the model. 

Given the lack of labels for the unlabelled pool, EGL algorithms define a distribution over the labels to compute the expected gradient. The formulation for EGL \cite{settles2009active, settles2008analysis} is shown in Eq: \ref{eq:og_egl}:
\begin{equation}
    x^{*} = \argmax_x \sum_i P(y_i | x; \theta)\, \| \nabla_{\theta} l\, (\mathcal{L} \cup \{x, y_i\}; \theta)\|
\label{eq:og_egl}
\end{equation}
Settles and Craven \cite{settles2009active} use the softmax outputs as a distribution over the labels $\bar{y}$, with the gradient computed assuming $y_i \in \bar{y}$ is the correct label for $x$. However, this approach is computationally expensive, and is also not representative of tasks where discrete probabilistic outputs are not available such as in regression.

\par
To overcome the lack of labels, Cai \etal \cite{egl_regression} extended expected gradient length to regression by building an ensemble of models. Let $\mathcal{F}^K = \{f_1, f_2 \ldots f_K\}$ represent a set of hypothesis obtained by training on subsets of labelled data $z = \{(x_1, y_1), (x_2, y_2) \ldots (x_N, y_n)\}$ and $f_z$ obtained by training on the entire pool of labelled data. Then, the sampling formulation can be represented as:
\begin{equation}
    x^{*} = \argmax_x \dfrac{1}{K} \sum_{k=1}^{K} \|(f_z(x) - f_k(x))x\|
\label{eq:regress_egl}
\end{equation}
Eq: \ref{eq:regress_egl} is intuitively defined and uses a committee of hypothesis to approximate the change induced by the sample $x$ in the original model $f_z$. Therefore, the samples picked for annotation are those where there is higher level of disagreement between the model $f_z$ and various weak learners $f_k$.


\section{Theory}

While Cai \etal \cite{batch_egl_regression, egl_regression} provides experimental evidence using Eq: \ref{eq:regress_egl}, we derive and show that Eq: \ref{eq:regress_egl} is a special case of the closed form solution obtained for EGL - regression. Further exploration reveals that the EGL framework too unifies aleatoric and epistemic uncertainty as done in \cite{kendall2017uncertainties}. Perhaps Bayesian uncertainty and EGL are two sides of the same coin?

\subsection{The No Free Lunch Theorem}

Let $\mathcal{Z} \in \{ \mathcal{X} \times \mathcal{Y}\}$ represent the domain and $z_{obs}$ denote the observed samples. We model the true distribution $p$ over $\mathcal{Z}$, parameterized by $\theta_0$ as $p(x, y | \theta_0) = p(y | x, \theta_0) \, p(x)$. Subsequently, \cite{huang2016active, ly2017tutorial, sourati2017asymptotic} define a new distribution $q$ for the observed values $z_{obs}$ as $q(x, y | \theta_0) = p(y | x, \theta_0) q(x)$ where $q(x)$ reflects the observed samples $x$ in $z_{obs}$. However, is it correct to assume that $\theta_0$ alone describes $z_{obs}$? 

\par
We argue that this assumption violates the No Free Lunch theorem \cite{shalev2014understanding}: Let $\mathcal{A}(z_{obs}) \rightarrow \theta$ represent the empirical risk minimizer (ERM) on the observed samples $z_{obs} \sim p_{z}$ which yields a hypothesis $\theta$ from the set of \textit{unbiased} hypotheses $\Theta$. Then for any hypothesis $\theta$ returned by the learner $\mathcal{A}$, there exists a distribution $p_z$ on which it fails. 

\par
A simpler, equivalent statement is: \textit{there is no universal learner, no learner can succeed on all learning tasks \cite{shalev2014understanding}}. This is based on two factors: 1) The ambiguity associated with the unknown distribution $p_z$ and 2) The set of unbiased hypotheses $\Theta$ not reflecting prior knowledge. Any learning algorithm $\mathcal{A}$ requires prior knowledge in the form of bias in the hypotheses $\Theta$. The lack of a bias indicates an infinite number of unique hypothesis that perfectly agree on $z_{obs}$, but disagree to varying degrees on the remaining sample space. Since we can only model $z_{obs}$ and not $p_z$, this implies any of the ERM hypotheses can be the true hypothesis, which forms the crux of the theorem. 

\par
The No Free Lunch theorem holds significance in the Active Learning paradigm. Active Learning assumes a small set of labelled samples $z_{obs} \sim p_z$, an initial trained model $\theta$ and a hypotheses class $\Theta$ (such as a neural network) with low bias. Since $z_{obs}$ does not adequately represent $p_z$ and the hypotheses set $\Theta$ has low bias relative to $z_{obs}$, the ERM learner potentially returns multiple $\theta$ with a low empirical risk as the initial trained model. Hence, while we can safely assume that $\theta_0$ parameterizes the true distribution $p$, the assumption that $\theta_0$ alone represents the observed samples is incorrect. Therefore, our distribution over the observed samples is in fact an expectation over all the multiple plausible hypotheses returned by an ERM learner that explain the observed samples $z_{obs}$.
\begin{align}
    q(x, y) & = \mathbb{E}_{\theta} \Big[ q(y | x, \theta) q(x) \Big] \nonumber \\
            & = \int_{\theta} q(y | x, \theta) q(x) \pi (\theta | z_{obs}) \mathrm{d} \theta
\label{eq:nfl}
\end{align}
\textit{Although representing the parameters as a distribution is standard practice in Bayesian analysis, the No Free Lunch theorem within the scope of active learning explains the existence of multiple hypothesis which can be quantified using bayesian analysis}.

\subsection{Fisher Information}

The proof proceeds further using a well known result on the asymptotic convergence of model parameters to its true value: $\sqrt{n}(\hat{\theta} - \theta^{*}) \xrightarrow{D} \mathcal{N}(0, I_q^{-1}(\theta^{*})))$. Related works \cite{huang2016active, settles2008analysis} argue that one way of converging to the true parameters (LHS) is by maximizing the Fisher Information with respect to the training distribution $q$. (minimizing the inverse). This is equivalent to maximizing $\mathbb{E}_{q(x, y)}[\nabla_{\theta}l(x, y, \hat{\theta}) \nabla_{\theta}^Tl(x, y, \hat{\theta})]$. Using Eq: \ref{eq:nfl}, we show that this is equivalent to maximizing:
\begin{align}
    q^{*} = \argmax_{q} \int_x  & q(x) \int_y \int_{\theta} q(y | x, \theta) \pi (\theta | z_{obs})  \nonumber \\ 
    & \| \nabla_{\theta_0} l(x, y, \theta_0) \|^2 \, \mathrm{d}\theta \, \mathrm{d}y \, \mathrm{d}x
    \label{eq:theta1}
\end{align}

Since integrating over the parameter space is intractable, we approximate $\pi(\theta | z_{obs})$ by bootstrapping and building an ensemble of learners $f_i$ parameterized by $\theta_i$,  where $z_i \subset z_{obs}$. Next, we define $f_z$ to be trained on all $z_{obs}$, with $\theta_z$ our best estimate of the true parameters $\theta_0$. We note that finding a training distribution $q$ that maximizes the expected gradient is same as including the sample $x$ having the highest expected gradient in the training set \cite{huang2016active}:
\begin{align}
    x^{*} = \argmax_x \dfrac{1}{K} \sum_{k=1}^{K}
    \int_y q(y | x, \theta_k) \| \nabla_{\theta_z} l(x, y, \theta_z) \|^2
    \label{eq:egl_pp_ensemble}
\end{align}

\subsection{Linear Regression}

Statistical linear regression results \cite{lec31pdf94:online} show that $q(y|x, \theta_k)$ is a gaussian distribution $\mathcal{N}(\mu_k(x), \sigma_k(x))$.

\begin{align}
    x^{*} & = \argmax_x \dfrac{1}{K} \sum_{k=1}^{K}
    \int_{-\infty}^{\infty} \mathcal{N}(y, \mu_k, \sigma_k) [(f_z(x) - y)x]^2 \, \mathrm{d}y
\label{eq:theta2}
\end{align}
Note that we have substituted $l(x, y, \theta_z)$ in Eq: \ref{eq:egl_pp_ensemble} with the regression mean square error for the model $f_z$. On comparing with Cai \etal (Eq: \ref{eq:regress_egl}), we note that the gradient term in our formulation is squared as a result of maximizing the Fisher Information. Also Eq: \ref{eq:regress_egl} is equivalent to Eq: \ref{eq:theta2}, if exactly one sample $y_i$ is drawn from  $q(y|x, \theta_i)$ for each $i \in K$ corresponding to the mode of the distribution.
\par
Eq: \ref{eq:theta2} has a closed form solution. Let $(y - f_z(x))^2 = [(y - \mu_k)) + (\mu_k - f_z)]^2$. The integral can be reduced to: $\int \mathcal{N}(y, \mu_k, \sigma_k)(y - \mu_k)^2\mathrm{d}y + 2\int \mathcal{N}(y, \mu_k, \sigma_k)(y - \mu_k)(\mu_k - f_z)\mathrm{d}y + \int \mathcal{N}(y, \mu_k, \sigma_k)(\mu_k - f_z)^2\mathrm{d}y$. The first term corresponds to the variance of the normal distribution, the second term results in a zero as the means cancel out and the last term coefficient is independent of $y$, resulting in $(\mu_k - f_z)^2$. The final closed form solution completing our derivation is:
\begin{equation}
    x^{*} = \argmax_{x \in \mathcal{U}} \dfrac{||x||^2}{K} \sum_{k=1}^{K} \sigma_k^2(x) + (\mu_k(x) - f_z(x))^2
    \label{eq:final}
\end{equation}

\subsection{Non-Linear Regression}
Bayesian linear regression allows us to model $q(y|x, \theta_i)$ as a Normal distribution, simplifying our analysis. Can we follow a similar approach to extend our analysis to non-linear regression? \\

\par
\noindent
\textbf{Aleatoric Uncertainty} \\
Regression assumes that the residuals are normally distributed $\epsilon \sim \mathcal{N}(0, \sigma)$. \cite{kendall2017uncertainties, nix1994estimating} show that if $\sigma$ is made learnable for each input, this reduces to computing the heteroscedastic aleatoric uncertainty for the input $x$. 
\begin{equation}
    l(x, y, \theta) = \dfrac{1}{2\sigma(x)^2} ||y - \hat{y} ||^2 + \dfrac{1}{2} \textrm{log}\, \sigma(x)^2
    \label{eq:aleatoric}
\end{equation}
If we let $\hat{y}, \sigma = f_{\theta}(x)$, Eq: \ref{eq:aleatoric} is a consequence of minimizing the negative log-likelihood over the normal distribution. \\

\noindent
\textbf{Gaussian Process} \\
Gaussian Process is a widely used statistical technique that imposes a distribution over the functions which can describe the observed data. The essence of Gaussian process is to represent each data point as a sample from a unique random variable, with the covariance between random variables being denoted by a kernel matrix $\mathcal{K}$. 
While a complete discussion is beyond the scope of the paper,  Gaussian Process regression allows us to model: $q(y|x, z_{obs}) \sim \mathcal{N}(\mu, \sigma)$ where $\mu, \sigma$ have a closed form solution \cite{cs229gau95:online}:
\begin{align}
    \mu & = \mathcal{K}(x, x_{obs}) \mathcal{K}(x_{obs}, x_{obs})^{-1} \bar{y}_{obs} \nonumber  \\ 
    \sigma & = \mathcal{K}(x, x) - \mathcal{K}(x, x_{obs}) \mathcal{K}(x_{obs}, x_{obs})^{-1} \mathcal{K}(x_{obs}, x)
\label{eq:gp}
\end{align}
Integrating Gaussian Process into the EGL framework is trivial. Instead of using parametric models $\theta_i$ to form an ensemble (Eq: \ref{eq:egl_pp_ensemble}), we can use an ensemble of Gaussian Process $q_i$ fitted on $z_i \subset z_{obs}$. \\

\noindent
\textbf{Formulation} \\
Both Aleatoric as well as Gaussian Process allow us to model $q(y|x, z_{obs}) \sim \mathcal{N}(\mu, \sigma)$. Also, the gradient for non-linear regression (from Eq: \ref{eq:egl_pp_ensemble}) can be split as: $\nabla_{\theta_z} l(x, y, \theta_z) = (y - f_z(x)) \nabla_{\theta_z} f_z(x)$, where $\nabla_{\theta_z} f_z(x)$ is independent of the ensemble. Hence, our analysis for linear holds true for non-linear regression too:
\begin{equation}
    x^{*} = \argmax_{x \in \mathcal{U}} \dfrac{||\nabla_{\theta_z} f_z(x)||^2}{K} \sum_{k=1}^{K} \hat{\sigma}_k^2(x) + (\mu_k(x) - f_z(x))^2
   \label{eq:egl_nonlinear}
\end{equation}

\subsection{Comparison with Predictive Uncertainty}

Kendall and Gal \cite{kendall2017uncertainties} define predictive variance as:
\begin{align}
    Var(y) \approx \dfrac{1}{K} \sum_{k=1}^{K} \hat{\sigma}_k^2 + \dfrac{1}{K} \sum_{k=1}^{K} f_k^2(x) - \bigg( \dfrac{1}{K} \sum_{k=1}^{K} f_k(x) \bigg)^2
\label{eq:kendall}
\end{align}
We highlight that both predictive uncertainty (Eq: \ref{eq:kendall}) and EGL (Eq: \ref{eq:final}, \ref{eq:egl_nonlinear}) use two measures of uncertainty. The first corresponds to local uncertainty ($\sigma_k$): what an individual model believes the uncertainty is. The second corresponds to global/epistemic uncertainty ($\mu - f$): measure of disagreement between models (from an ensemble/dropout).

If we consider Aleatoric - EGL, then approximating $ \| \nabla_{\theta_z} l(x, y, \theta_z) \| \approx c\| (y - f_z(x)) \| $ allows us to derive predictive uncertainty from EGL. The basis of this approximation is that \textit{chain rule} ensures that the gradient is highly correlated with the residual. 

\subsection{Significance}
Although Cai \etal \cite{batch_egl_regression, egl_regression} provided experimental evidence for expected length in regression, their formulation was intuitively driven. Our theory supports this experimental evidence, highlighting that the original formulation (Eq: \ref{eq:regress_egl}) is a special case of our derived formulation. We then proceed to show that Expected Gradient Length for regression can unify both aleatoric and epistemic uncertainty, proving equivalence with \cite{kendall2017uncertainties}.

\section{Algorithm: EGL++}

Our theory shares the following assumptions with bayesian uncertainty: 1) The underlying task supports ensembles/dropouts 2) $q(y|x,\theta)$ can be determined analytically. While our assumptions hold true for most regression tasks, can we adopt EGL for active learning to tasks where our assumptions are violated? Our discussion in Sec: \ref{par:challenge} highlights our inability to use dropouts or ensembles with human pose estimation (HPE). Additionally, modelling $q(y|x,\theta)$ as aleatoric uncertainty is tedious since HPE models regress 2-D heatmaps.

\par
To circumvent these challenges, we propose EGL++, an alternative interpretation of the EGL framework. EGL++ approximates the effect of ensembles by leveraging previously observed labels $y_j$ as potential labels for unlabelled images $x_i$.  We facilitate this by representing each sample as an \textit{image-embedding-(label/prediction)} triplet, and use t-SNE to quantify the neighborhood in the embedding space. This approach makes EGL++ highly interpretable, since it quantifies our intuition that similar inputs have similar representations and thus predictions. 

Therefore, in the absence of a closed form solution, EGL++ solves:
\begin{align}
    x^{*} = \argmax_{x \in \mathcal{U}} \sum_{n=1}^{n=N} q_{tsne}(y_n | x, \theta) \| \nabla_{\theta} l(x, y_n, \theta) \|^2
    \label{eq:egl_pp1}
\end{align}

We discuss each of the components: $q_{tsne}(y|x, \theta)$ and $\| \nabla_{\theta} l(x, y_n, \theta) \|^2$ in greater detail: \\

\begin{figure*}
\begin{center}
\includegraphics[width=0.9\linewidth]{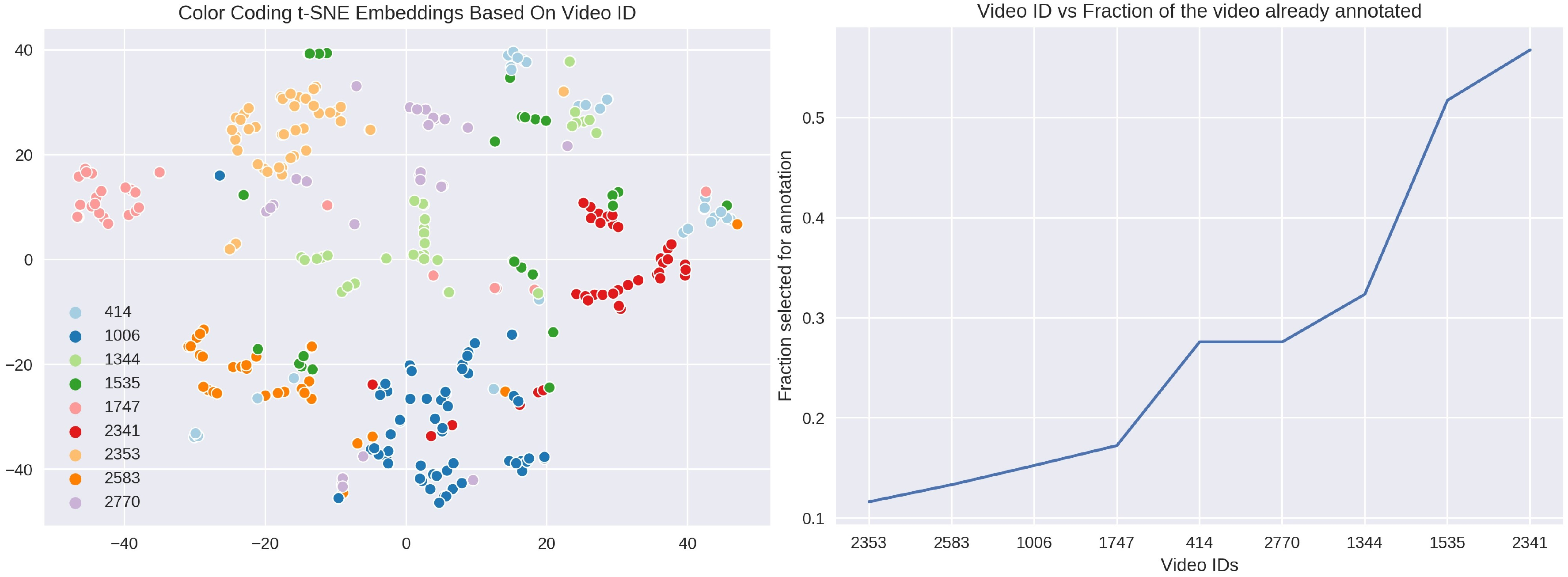}
\vspace*{-1.25em}
\end{center}
   \caption{(Left) t-SNE visualization of MPII, color-coded by video. (Right) Fraction of frames chosen per video. Heavily sampled videos (1344, 1535, 2341) either have multiple clusters across the embedding space or have overlapping representations with other videos. In contrast, lightly sampled videos (2353, 2583) have embeddings concentrated in a small space, needing fewer labelled samples.} 

\label{fig:tsne}
\end{figure*}

\par
\noindent
\textbf{Conditional Probability} \\
Computing $q(y|x, \theta)$ can be further split into computing network representations $h$ for the set of images $x$, and modelling the relation between these representations $h$. Introducing network representations has two major benefits. Not only do smaller representations improve compute time, network embeddings allow us to model semantic relations established by the neural network.

Dimensionality reduction algorithms are useful in modelling the spatial representation between high dimensional samples. While classical algorithms such as Locally Linear Embedding \cite{roweis2000nonlinear} beautifully describe the sample space, they have high time complexity O$(n^3)$ due to eigen-decomposition. Hence, we turn our attention to t-SNE, a popular visualization algorithm with O$(n^2)$ time complexity and optimized for GPU usage.

\textbf{t-SNE}: t-SNE \cite{hinton2002stochastic, let-sne, JMLR:v15:vandermaaten14a, tsne} provides a convenient method to model $q(y | x, \theta)$ in high-dimensional space.
Let $h_i$ represent the embedding for the $i^{th}$ sample. Then the probability of $h_j$ being a neighbor of $h_i$ is:

\begin{equation}
    q(h_j | h_i, \theta) = \dfrac{exp( - \| h_i - h_j\|^2 / 2\sigma_i^2)}
    {\sum_{k \ne i} exp( - \| h_i - h_k\|^2 / 2\sigma_i^2})
\label{eq:p_ji}
\end{equation}

The neighborhood for every sample $i$ is quantified using a gaussian distribution centered at $h_i$ with variance $\sigma^2_i$. Perplexity \cite{tsne} tunes the variance to account for variable density in the embedding space, thereby determining the spatial extent of the neighborhood for $h_i$. 

Since we know the ground truth $y_j$ for each embedding $h_j$ when image $x_j$ is from the labelled set, our heuristic lies in approximating $q(h_j | h_i, \theta)$ with $\hat{q}(y_j | h_i, \theta)$. This heuristic allows us to leverage previously annotated data and quantify the probability of $y_j$ being the true label for $x_i$. We divide our embeddings into labelled $h_{\mathcal{L}}$ and unlabelled sets $h_{\mathcal{U}}$ where we identify the neighborhood for each $h \in \mathcal{U}$ in terms of all $h \in \mathcal{L}$. This results in a conditional probability matrix $q(h_{\mathcal{L}} | h_{\mathcal{U}}, \theta) = \hat{q}(y_{\mathcal{L}} | h_{\mathcal{U}}, \theta)$ of shape $(|\mathcal{U}| \times |\mathcal{L}|)$, allowing us to compute the expectation for $\| \nabla_{\theta} l(x_{\mathcal{U}}, y_{\mathcal{L}}, \theta) \|^2$. \\

\par
\noindent
\textbf{Gradient Computation} \\
In a utopian world where deep learning models train instantaneously, gradient computations would nary have been an issue. Alas, that is not the case, and EGL++ needs fast yet efficient heuristics for gradient computation. We use two approximations to speed up EGL++. The first approximation is in limiting the number of neighbours per sample. Previously we had computed a conditional probability matrix $\hat{q}(y_j | h_i, \theta)$, which is used in calculating the expected gradient length $\sum_j \hat{q}(y_j | h_i, \theta) \| \nabla_{\theta} l(x_i, y_j, \theta) \|^2$ for the unlabelled sample $x_i$. Computing this value for all labels $y_j$ is an expensive task, especially when most values of $\hat{q}(y_j | h_i, \theta) \approx 0$. Instead, we restrict $y_j$ to be among the top-K most probable values given by $\hat{q}(y_j | h_i, \theta)$, limiting the number of gradient computations needed.

\par
The second approximation uses Goodfellow's technique for efficient per example gradient computation \cite{efficient}. Computing gradients per example is difficult since popular deep learning frameworks aggregate gradients across the minibatch. Goodfellow's solution uses intermediate representations to compute the gradient in convolutional and linear layers for each sample $x_i$ in the mini-batch . While this is not the true gradient for $x$, (for \eg, no batch normalization gradients) this heuristic allows for fast computations of gradients in a single forward pass.

\begin{figure*}[h]
\begin{center}
\includegraphics[width=0.7\linewidth]{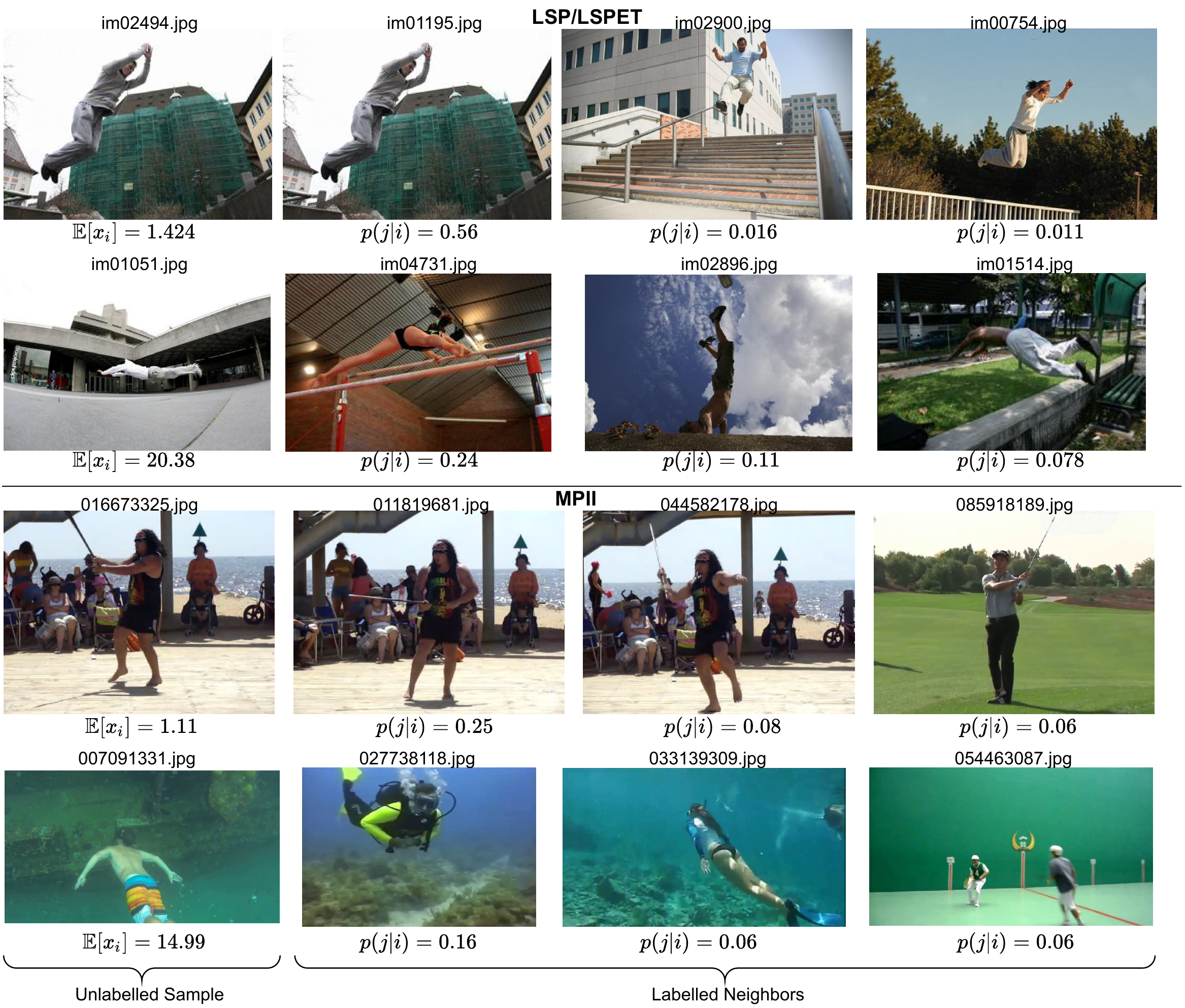}
\vspace*{-1.25em}
\end{center}
   \caption{\textit{Interpretability}: EGL++ quantifies the intuition that similar inputs have similar representations. Unlabelled images (first column) with the smallest expected gradient already have similar annotated images. In contrast, unlabelled images with the highest expected gradient share limited semantic similarity with its nearest neighbours. \textit{(More figures in the supplementary material)}}
\label{fig:nbrs}
\end{figure*}

\begin{algorithm}
    \label{algo:1}

\DontPrintSemicolon
  
  \KwInput{Human pose estimator ($f$), Number of neighbors ($\mathcal{K}$), Budget ($\mathcal{B}$)}
  \KwOutput{Unlabelled images $(x^{*}_{\mathcal{U}})$ for annotation}
  \KwData{Labelled ($x_\mathcal{L}-y_{\mathcal{L}}$) and Unlabelled ($x_\mathcal{U}$) images}
  
  $h_{\mathcal{U}} = f_{enc}(x_\mathcal{U})$  \tcp*{encodings for unlabelled}
  $h_{\mathcal{L}} = f_{enc}(x_\mathcal{L})$  \tcp*{encodings for labelled}
  Compute matrix $q(h_j | h_i, \theta)$ using Eq: \ref{eq:p_ji}, where $i \in \mathcal{U}$ and $j \in \mathcal{L}$ \tcp*{Size ($|\mathcal{U}| \times |\mathcal{L}|$)} 
  $\hat{q}(y_j | h_i, \theta) = q(h_j | h_i, \theta)$ \tcp*{Heuristic, $h_j \rightarrow y_j$}
  Initialize $\{\theta_{fast}: \theta_0 \textrm{layers} \in \textrm{linear, conv}\}$ \\
  $(\hat{q}(y_j | h_i, \theta), y_j)$  = top - $\mathcal{K}(\hat{q}(y_j | h_i, \theta), y_j)$ \\ 
  \tcc{New $\hat{q}$, $\nabla$ matrix size: $|\mathcal{U}| \times |\mathcal{K}|$}
  \tcc{CPU multiprocessing - parallelize \textit{for}}
  \For{$x_i$ in $\mathcal{U}$}    
        {   
            \tcc{GPU parallel across $j$ for each $i$}
            \For{$y_j$ in $\mathcal{L}$}
            {
                \tcc{Fast Gradient Approximation}
                Compute gradient norm $\| \nabla_{\theta_{fast}} f(x_i), y_j \|^2$
            }
        }
  
  \tcc{Expectation is hadamard product $\hat{q} \circ \nabla$, reduce summed along $j$. Shape: $|\mathcal{U}|$}
  $\bar{\mathbb{E}}_{y}[\nabla_{\theta}(y, x_{\mathcal{U}}, \theta)] = \sum_j \hat{q}(y_j | h_i, \theta) \circ \| \nabla_{\theta_{fast}} f(x_i), y_j \|^2$ 
  
  \Return $x^{*}_{\mathcal{U}}$: Return samples corresponding to  top - $\mathcal{B}$ expected gradients from $\bar{\mathbb{E}}_{y}[\nabla_{\theta}(y, x_{\mathcal{U}}, \theta)]$
  
\caption{EGL++}
\end{algorithm}

\section{Experiments and Results}

\textbf{Goal:} EGL and Bayesian uncertainty provide theoretical results, but what happens when the underlying assumptions aren't feasible? We turn the spotlight on active learning for human pose estimation (HPE), with our previous discussion summarizing the associated challenges with the architecture (Sec: \ref{par:challenge}).
While EGL++ extends the EGL paradigm to human pose, can we achieve competitive performance in comparison to popular active learning algorithms for HPE? \textit{Keeping in mind practical applications such as an incoming data stream, can EGL++ maximize the model performance per set of images annotated?
}
\par
\textbf{Experimental Design}:
Our code (in PyTorch \cite{NEURIPS2019_9015}) is available on \url{https://github.com/meghshukla/ActiveLearningForHumanPose}. We compare EGL++ with three state-of-the-art algorithms in active learning for human pose: Coreset \cite{sener2017active}, Multi-peak entropy \cite{liu2017active} and Learning Loss \cite{learnloss}. We simulate multiple cycles of active learning, with each cycle selecting 1000 new images for annotation. Base models trained on an initial random 1000 images are shared by all algorithms. Following \cite{liu2017active, learnloss}, we use MPII \cite{mpii} and LSP/LSPET \cite{lsp, lspet} human pose datasets. The MPII dataset consists of images corresponding to everyday human activity, whereas LSP emphasizes on sporting activities. For MPII, we report our results on the Newell validation split \cite{hg}, as done in \cite{liu2017active, learnloss}. We use the LSP authors' data split; the first 1000 images used for training the base models, and the last 1000 images acting as the testing set. The entire LSPET dataset is used as the unlabelled pool for active learning sampling. We use the smaller two-stacked Hourglass \cite{hg} with PCKh of 88\% (following \cite{learnloss}) as the human pose estimator. Following Multi-Peak Entropy, we extract single persons into separate images if the underlying image contains multiple persons. We also use standard evaluation metrics: PCKh@0.5 for MPII and PCK@0.2 for LSP-LSPET.

\begin{table*}
\footnotesize
\centering
\resizebox{2\columnwidth}{!}{%
\renewcommand{\arraystretch}{0.6}
\begin{tabular}{llll|lll|lll|lll|lll}

&  \multicolumn{14}{c}{MPII Newell Validation Split: Mean+-Sigma (5 runs), one-tailed paired t-test (vs EGL++) at 0.1 significance value} \\
\toprule
\#images $\rightarrow$  &  \multicolumn{3}{c|}{2000} & \multicolumn{3}{c|}{3000} & \multicolumn{3}{c|}{4000} & \multicolumn{3}{c|}{5000} & \multicolumn{3}{c}{6000}\\
\midrule
Methods & $\mu$ & $\sigma$ & $p-value$ & $\mu$ & $\sigma$ & $p$ & $\mu$ & $\sigma$ & $p$ & $\mu$ & $\sigma$ & $p$ & $\mu$ & $\sigma$ & $p$  \\ 

\midrule
Random & 75.95 & 0.55 & \textbf{0.003} & 78.33 & 0.65 & \textbf{0.012} & 80.31 & 0.91 & \textbf{0.006} & 81.35 & 0.41 & \textbf{0.001} & 82.23 & 0.74 & \textbf{0.007} \\

Core-set \cite{sener2017active} & 76.61 & 0.6 & \textbf{0.0047} & 79.24 & 0.7 & 0.245 & 81.25 & 0.67 & \textbf{0.072} & 82.23 & 1.14 & 0.123 & 82.97 & 1.11 & \textbf{0.064} \\

Multi-peak \cite{liu2017active} & 76.74 & 0.61 & \textbf{0.0054} & 79.56 & 0.46 & 0.462 & 81.19 & 0.31 & \textbf{0.063} & 82.61 & 0.5 & \textbf{0.093} & 83.11 & 0.71 & \textbf{0.062} \\

Learning Loss \cite{learnloss} & 76.28 & 0.76 & \textbf{0.0276} & 79.27 & 0.52 & 0.185 & 81.35 & 0.35 & 0.152 & 82.94 & 0.44 & 0.319 & 83.79 & 0.46 & \textbf{0.053} \\

EGL++ (ours) & \textit{77.28} & 0.63 & - & \textit{79.58} & 0.33 & - & \textit{81.53} & 0.51 & - & \textit{83.07} & 0.25 & - & \textit{84} & 0.38 & - \\

\bottomrule
&  \multicolumn{15}{c}{\Large } \\

&  \multicolumn{14}{c}{LSP Test Split: Mean+-Sigma (5 runs), one-tailed paired t-test (vs EGL++) at 0.1 significance value} \\
\toprule
\#images $\rightarrow$  &  \multicolumn{3}{c|}{2000} & \multicolumn{3}{c|}{3000} & \multicolumn{3}{c|}{4000} & \multicolumn{3}{c|}{5000} & \multicolumn{3}{c}{6000}\\
\midrule
Methods & $\mu$ & $\sigma$ & $p-value$ & $\mu$ & $\sigma$ & $p$ & $\mu$ & $\sigma$ & $p$ & $\mu$ & $\sigma$ & $p$ & $\mu$ & $\sigma$ & $p$  \\ 

\midrule
Random & 80.34 & 0.31 & 0.285 & 81.81 & 0.24 & 0.297 & 82.68 & 0.32 & 0.138 & 83.35 & 0.36 & \textbf{0.095} & 84.13 & 0.16 & \textbf{0.029} \\

Core-set \cite{sener2017active} & 79.69 & 0.82 & \textbf{0.043} & 81.41 & 0.45 & \textbf{0.096} & 82.25 & 0.39 & \textbf{0.021} & 83.11 & 0.38 & \textbf{0.032} & 83.73 & 0.31 & \textbf{0.006} \\

Multi-peak \cite{liu2017active} & 80.36 & 0.4 & 0.225 & 81.48 & 0.53 & 0.125 & 82.63 & 0.23 & 0.119 & 83.29 & 0.2 & \textbf{0.036} & 84.3 & 0.44 & \textbf{0.063} \\

Learning Loss \cite{learnloss} & 79.58 & 0.39 & \textbf{0.002} & 81.39 & 0.34 & \textbf{0.071} & 82.31 & 0.42 & \textbf{0.038} & 83.31 & 0.25 & \textbf{0.029} & 84.2 & 0.53 & \textbf{0.098} \\

EGL++ (ours) & \textit{80.49} & 0.45 & - & \textit{81.91} & 0.27 & - & \textit{83.03} & 0.43 & - & \textit{83.91} & 0.51 & - & \textit{84.68} & 0.36 & - \\
\bottomrule
\vspace*{-1.5mm}
&  \multicolumn{10}{c}{\Large } \\
%
%
\end{tabular}
}
\caption{(Active Learning Simulation) While \cite{liu2017active} establishes active learning for human pose, we focus on a more practical challenge: maximizing model performance per set of images annotated. The quantitative results and p-value from t-test for significance indicate that EGL++ is competitive, performing equivalent to if not better than its peers across all active learning cycles.}
\label{tab:res}
\vspace*{-1.5mm}
\end{table*}

\subsection{Qualitative Assessment} 

Fig: \ref{fig:tsne} confirms our hypothesis that similar inputs have similar representations. Videos with multiple clusters across the embedding space usually exhibit high variation in content, therefore requiring a higher number of selected frames to represent the different clusters. Similarly, videos with overlapping representations have a higher rate of sampling. Videos with concentrated grouping of embeddings indicate content with limited movements, and can be represented with fewer labelled samples.

Fig: \ref{fig:nbrs} highlights the interpretability associated with EGL++. Images with the lowest EGL score  are well represented in the training set, leaving little ambiguity in the potential label. In contrast, images with the highest EGL score do not share semantic similarities with its neighbours, leaving considerable ambiguity over the potential label. We also note that EGL++ fairs well in interpretability compared to its peers. Learning Loss and its variants predict the 'loss' for an image but lacks explanation. Core-set encourages diversity in the sampled set, but fails to explain the selection of individual images in the set. Multi-peak entropy has higher explainability since it captures the ambiguity within the heatmap, however the approach does not reason for the ambiguity given an image.

\subsection{Quantitative Assessment}

\textbf{Test for statistical significance: } We report the mean ($\mu$), std. dev. ($\sigma$) and \textbf{\textit{p-value}}, the latter obtained from the paired t-test to measure the significance of our results at a 90\% confidence level \cite{TestsofS11:online}. The paired t-test is advantageous since: 1) It accounts for random good/bad initializations of the base models ( mean by definition averages it) 2) It is independent of the number of samples selected per cycle. Since single person datasets have limited training examples (MPII: \httilde{}20K, LSP-LSPET: \httilde{}11K), the number of samples selected per cycle is small. This directly translates into smaller jumps in mean across all methods, as is evident from Tab: \ref{tab:res}.

\textbf{Comparison: } MPII and LSP-LSPET allow us to examine the behavior of the algorithms under two different conditions: when the labelled and unlabelled pool follow the same distribution (MPII), and when the pools differ (LSP - initial labelled, LSPET - unlabelled). While active learning algorithms perform significantly better (based on p-value) than random sampling on MPII, these algorithms struggle in the initial cycles of LSP-LSPET due to the change from LSP to LSPET. Also, learning loss being a deep learning based method improves in performance as the number of labelled instances increase. We also observe that Core-Set does not extend well to human pose, since linear separability of embeddings from the penultimate layer is not enforced by human pose estimation. Multi-peak entropy fares well in the initial phases of active learning, with its performance declining in the subsequent stages. Entropy for active learning has been well studied in \cite{learnloss}, and while multi-peak entropy fares better than vanilla entropy, it shares the same weaknesses as its predecessor. Like its peers, EGL++ too struggles to outperform random sampling when the model shifts from LSP to LSPET. However, EGL++ is the fastest to recover, performing better than its peers based on the p-value. In general, we observe that at the minimum, EGL performs as good as the best active learning technique. 

\section{Conclusion}

This paper explores expected gradient length (EGL) in regression from a theoretical perspective, making two key contributions. The first contribution derives the closed form solution for EGL-regression, thereby supporting the experimental evidence in literature. Specifically, we show that the intuitive formulation in literature is in fact a special case of our derived solution. \textit{Our second contribution lies in highlighting that EGL-regression unifies both: aleatoric and epistemic uncertainty, allowing us to draw parallels with predictive uncertainty.} With EGL++, we adopt the EGL framework to tasks where creating ensembles or computing a distribution over the labels is infeasible. EGL++ approximates the effect of ensembles by quantifying the probability of existing labels being the true label for an unlabelled image. This approximation allows us to extend the EGL framework to human pose estimation. Our experiments validate that EGL++ is interpretable and competitive in comparison to popular active learning algorithms for human pose estimation.

\vspace*{\fill}

\noindent
\textit{\textbf{Acknowledgement}: I thank Brijesh Pillai and Partha Bhattacharya, MBRDI, for providing funding for this work.}

\newpage

{\small
\bibliographystyle{ieee_fullname}
\bibliography{egbib}
}

\end{document}